\def\year{2020}\relax
\documentclass[letterpaper]{article} 
\usepackage{aaai20}  
\usepackage{times}  
\usepackage{helvet} 
\usepackage{courier}  
\usepackage[hyphens]{url}  
\usepackage{graphicx} 
\usepackage{xcolor}
\usepackage{tabularx}
\usepackage{hyperref}
\usepackage{graphicx}  
\title{\textsc{ManyModalQA}: Modality Disambiguation and QA \\ over Diverse Inputs}

\author{Darryl Hannan, Akshay Jain, and Mohit Bansal\\
University of North Carolina at Chapel Hill\\
{dhannan, akshayj, mbansal}@cs.unc.edu}

\newcommand{\citep}{\cite}
\newcommand{\citet}[1]
{\citeauthor{#1} \shortcite{#1}}

\def\modelname{\textsc{ManyModalQA}}

\begin{document}

\maketitle

\begin{abstract}
We present a new multimodal question answering challenge, \modelname{}, in which an agent must answer a question by considering three distinct modalities: text, images, and tables. We collect our data by scraping Wikipedia and then utilize crowdsourcing to collect question-answer pairs. Our questions are ambiguous, in that the modality that contains the answer is not easily determined based solely upon the question. To demonstrate this ambiguity, we construct a modality selector (or disambiguator) network, and this model gets substantially lower accuracy on our challenge set, compared to existing datasets, indicating that our questions are more ambiguous. By analyzing this model, we investigate which words in the question are indicative of the modality. Next, we construct a simple baseline \modelname{} model, which, based on the prediction from the modality selector, fires a corresponding pre-trained state-of-the-art unimodal QA model. We focus on providing the community with a new manymodal evaluation set and only provide a fine-tuning set, with the expectation that existing datasets and approaches will be transferred for most of the training, to encourage low-resource generalization without large, monolithic training sets for each new task. There is a significant gap between our baseline models and human performance; therefore, we hope that this challenge encourages research in end-to-end modality disambiguation and multimodal QA models, as well as transfer learning.
\end{abstract}

\begin{figure*}
    \centering
    \includegraphics[width=0.85\textwidth]{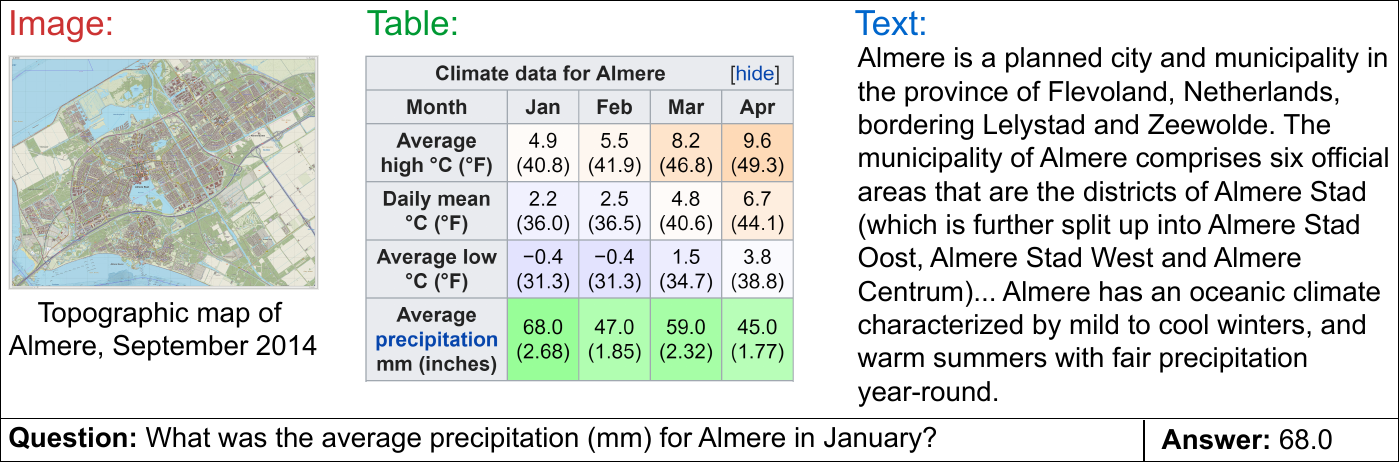}
    \caption{A condensed example of our task, using a sample that we collected from a Wikipedia page.}
    \label{fig:full_example}
\end{figure*}

\section{Introduction}

In real-world QA scenarios, questions might be answerable via a diverse set of resources such as documents, images, instructional videos, online reviews, data tables, diagrams, etc., any number of which could contain the answer. 
Wikipedia is an example of one such information source. Users turn to Wikipedia to answer many different inquiries and depending on the nature of the question, one modality may be better suited for answering the question than another. For example, consider the English Wikipedia page for the United States, which contains text, tables, and images. If a user has a question about the US census, numerical inquiries such as this are frequently answered by tables, and census information is in fact contained in a table on the page. Alternatively, the question ``When was the Declaration of Independence signed?" is unlikely to be answered by an image or table; users will naturally look at the text to answer this question. Amazon product pages are another real-world manymodal information source. When consumers have inquiries about a given product, they look at the provided images and videos, the textual description of the product, data tables containing the specifications, and reviews left by previous buyers (which can contain images or text), to locate the answer.
In contrast to this manymodal\footnote{We use the term ``manymodal'' in contrast to ``multimodal'' to emphasize that there are more than 2 modalities, contrasting what has typically been done by the community. In future work, this term may encompass more than the modalities that we consider here, potentially including video, audio, and more.} reality, the research community has been primarily operating in a unimodal QA setting, with some approaches involving two modalities such as VQA \citep{VQA}. By moving towards manymodal QA, models acquire access to different kinds of information that is not easily expressed in text, such as spatial and temporal information in images and videos. In this work, we take steps towards bridging the gap between real-world QA and the community's QA efforts by introducing a new manymodal QA challenge\footnote{We include some data samples in our appendix and the challenge set is available at 
\url{https://github.com/hannandarryl/ManyModalQA}.}, \modelname{}, that spans text, images, and tables, and explores the interaction between natural language and these other modalities. Our challenge set contains 2036 fine-tuning, 3055 dev, and 5099 test examples, providing enough examples for fine-tuning and a large test set to accurately measure performance. Each example contains textual information along with a table, an image, or both. Examples span a wide variety of knowledge domains, including politics, science, sports, and more.

In moving to this new manymodal setting, we prioritize maximizing the utility of prior unimodal and bimodal datasets and approaches,
as opposed to producing a large standalone dataset, which often leads to overfitting and over-tuning.
Our challenge specifically encourages the development of low-resource generalization learning techniques that utilize several large unimodal/bimodal tasks to complete a joint manymodal task with only enough data for fine-tuning.
We consider manymodal QA to consist of two sub-problems. First, the modality that is best used to answer the question must be determined. Then, this selected modality can be used to answer the question.\footnote{These steps can be explicitly separated (as in our baseline method) or implicitly performed in a future end-to-end model.} Hence, to make this task challenging (similar to real-world examples), we focus on increasing the difficulty of modality selection by making questions ambiguous with respect to which modality should be used to answer the question.
We encourage this ambiguity during data collection by framing our question collection crowdsourcing task as an adversarially oriented game, in which crowdworkers are tasked with creating a question, for a given modality, in a way that is ambiguous and will fool a QA robot designed for our task.
To verify the ambiguity of our questions, we also construct a modality selection network that, given the question, attempts to predict the modality that the answer will be found in. We use this model to quantitatively compare our data to data created by simply concatenating existing unimodal datasets and demonstrate that ours is more ambiguous.
We also present saliency visualizations to investigate which words frequently correspond to each modality channel, providing insight into which question words are used to predict each modality.

Our modality selector is then used to construct a baseline \modelname{} model, namely the `modality selection and question-answering network', which establishes initial results on our data. Our network combines the modality selector with 3 pre-trained state-of-the-art QA models on 3 popular unimodal datasets (with simple modifications to make them more uniform and better match our data). The question is first fed into the modality selector, then the pre-trained QA model corresponding to the predicted modality is run to answer the question.
This final architecture obtains 39.73\% on our test data, while the upper-limit human performance is 91.58\%, demonstrating that a substantial amount of future work remains on our new challenge set.

Overall, \modelname{} makes four primary contributions to the community:
\begin{enumerate}
    \item It provides the community with a new manymodal QA challenge, requiring the development of important techniques in adapting existing unimodal systems in Text-QA, Image-QA, and Table-QA, for manymodal evaluation.
    \item It considers 3 distinct modalities (text, images, and tables), which is especially important because tables are underexplored within the context of question answering.
    \item It is the first task to focus on modality disambiguation within the context of QA, encouraging the development of QA models that are better suited for real-world QA scenarios where the question's answer modality is unknown.
    \item The examples in our challenge set span a variety of knowledge domains, containing information about people, places, science, and more, in contrast to existing multimodal QA datasets.
\end{enumerate}

\section{Related Work} \label{sec:rel_work}
\subsection{Multimodal QA Datasets}
There are many tasks in which the input is an image and question, and the output is a textual answer~\citep{balanced_vqa_v2,clevr,Maharaj2017ADA,Kafle2018DVQAUD}. These tasks are multimodal, but the text component is minimal. We are interested in a stronger form of multimodality, which requires the agent to not just translate from one modality to another, but to learn to combine them, for example~\citep{lei2018tvqa,Rohrbach2015ADF,RobertLLogan2017MultimodalAE}.
Despite some prior work possessing this stronger sense of multimodality, most still only consider two modalities, text and images. We extend this by adding tabular information. Furthermore, we maximize the ambiguity of our questions, forcing a good model to consider each modality to determine the answer.

Textbook QA (TQA) \citep{textbookqa}, RecipeQA \citep{recipeqa} and TVQA \citep{lei2018tvqa} are the closest existing works to \modelname{}. TQA and RecipeQA require an agent to consider text and images to answer questions, while TVQA requires an agent to consider video and text subtitles to answer questions. The most obvious difference between these works and ours is that we include tables in our challenge, increasing the ambiguity of the questions and encouraging research in an important, under-investigated domain. Moreover, we specifically focus on generating questions that are representative of real-world QA, where the question's answer modality is unknown, forcing the model to consider more than one modality before answering the question. While TQA asks questions about middle school science, RecipeQA about recipes, and TVQA about 6 TV shows, our data is from Wikipedia, which is more open-ended, yielding questions about places, science, languages, and more. Additionally, TQA, RecipeQA, and TVQA are all multiple choice, whereas our answers are open-ended, making the task more challenging and leading to more robust models. Additionally, TVQA questions are compositional questions that follow a particular sentence structure, and RecipeQA contains automatically generated question-answer pairs, unlike our data that contains fully open-ended natural questions made by humans. Furthermore, we structure our task in a way that encourages low-resource generalization, mitigating the issue of over-fitting and over-tuning on the training set, which is commonly seen in large, stand-alone datasets.

\subsection{Unimodal Text and Table QA Datasets}
Text-based QA datasets exist in a variety of forms \citep{msmarco,raceqa,narrativeqa,squad20,coqa}. The introduction of the Stanford Question Answering Dataset (SQuAD) \citep{Rajpurkar2016SQuAD10}, has made text-based QA a popular task, leading to the development of several new approaches. CoQA \citep{coqa} is similar to traditional text-based QA, except the context document takes the form of a dialogue instead of a story or article. Alternatively, QAngaroo and HotpotQA~\citep{wikihop,yang2018hotpotqa} expand upon text-based QA by requiring agents to combine information from multiple documents, a process known as multi-hop reasoning.

Table-based QA datasets are fewer and the task is not as well-studied; a few large datasets are available, such as WikiTableQuestions \citep{wikitables}, TabMCQ \citep{Jauhar2016TablesAS}, and the MLB dataset \citep{cho2018adversarial}. WikiTableQuestions is the closest to our table examples because it has natural language questions and is collected from Wikipedia. TabMCQ is multiple choice, and the answers are always found in the table, whereas the tables in our challenge are all open-ended answers. The MLB dataset is similar in size to WikiTableQuestions, yet the domain is restricted as it only considers baseball tables and the questions are produced from templates. \citet{SunCellSearch} considers a different form of table-QA, where there is a large set of tables that must be searched to answer the question. In our task, only a single table is provided at any given time, making it closer to the previously mentioned datasets.

\section{The Task: ManyModal QA Reasoning} \label{sec:task_desc}
 Our task requires an agent to consider up to three distinct modalities (images, tables, and text), disambiguate a question, and produce an answer. Each example consists of a simplified Wikipedia page and contains at least two of these modalities. The question has been constructed to be ambiguous with respect to the modality that contains the answer.\footnote{The answer is always guaranteed to be found in at least one of the modalities but may be found in multiple modalities. All answers are a single word because this makes it easy to define a loss function and evaluate a model's performance. It also reduces noise during data collection as it limits the number of possible ways that the same answer can be expressed. In many cases, even datasets that consist of multi-word answers are still restricted to single entities and/or the average answer length is just a few words \citep{cnndailymail,searchqa}.}
 
 \noindent{\bf{Question Ambiguity}} \label{sec:ambiguity_desc}
 Ambiguity is critical to our challenge because it is the primary means through which we ensure that the task is manymodal, similar to real-world scenarios. We do not want the modality that contains the answer to be obvious based upon the question. For instance, ``According to the table, what was the average precipitation for Almere in January?" is not ambiguous because the question contains the word ``table", whereas the question in Figure~\ref{fig:full_example} is ambiguous because it could be found in any modality. If this ambiguity did not exist, the task would simply be a combination of unimodal tasks. The ambiguity forces the model to consider each of the modalities before determining which contains the answer, thereby increasing the difficulty and making it more representative of real-world QA. Making questions completely ambiguous is unreasonable and will result in questions that are unnatural; image questions will always be more visually oriented, text questions more factual, etc. However, we strive to make the questions as ambiguous as possible, while maintaining their quality.\footnote{We experimented with forcing questions to be strictly multimodal, i.e., the question can \emph{only} be answered by combining information from all three modalities. However, we found that this is extremely challenging for crowdworkers.}
 
 \noindent{\bf{QA Model Transfer Setup}}
Models are only as good as the datasets that they are trained on and it is challenging to construct a large training set that accurately represents the problem of manymodal QA.
There already exist high-quality, unimodal and bimodal, publicly available QA datasets, which have been improved over multiple releases.
Therefore, instead of collecting our own data from scratch, introducing new biases, we build upon prior work by only providing a fine-tuning set, while instead focusing on providing a high-quality evaluate set. This allows us to focus on the manymodal and ambiguity aspects, improving the quality of our data and promoting research in transfer learning. 
This setup encourages research in optimally combining multiple datasets and model architectures. Most work done in machine learning involves training and evaluating on a single dataset, leading to overfitting, over-tuning, and over-optimizing. Researchers may perform this procedure on multiple datasets to verify their results, but each is considered separately. Our challenge  instead encourages research in low-resource generalization from models trained on unimodal sub-tasks to a single manymodal task.

\section{Data Collection}

\noindent{\bf{Data Source}}
We collect our challenge data from English Wikipedia, which contains 6 million articles, each containing many modalities, including text, tables, images, video, audio, and more. Furthermore, all of this content is publicly available and easy to access.

\begin{figure}[t]
\centering
\includegraphics[width=0.4\textwidth]{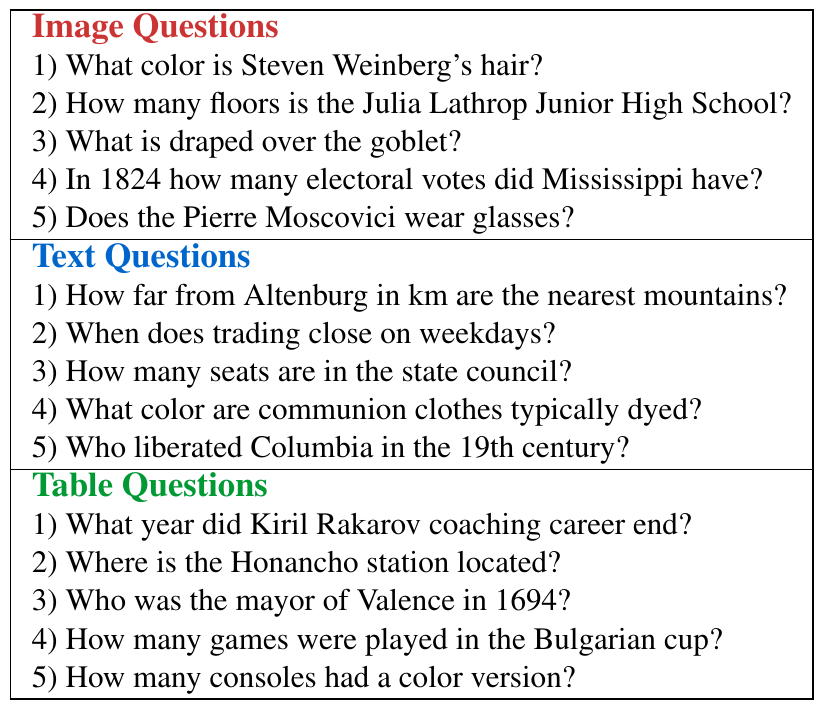}
\caption{Sample questions collected for each modality.}
  \label{tab:sample_questions}
 \end{figure}

\noindent{\bf{Crowdsourcing}} \label{crowdsourcing}
Wikipedia provides us with plenty of raw information but does not provide questions. While questions can be automatically generated, and this has been done in prior work \citep{recipeqa,clevr}, we have human annotators create questions via Amazon Mechanical Turk (MTurk), leading to diverse, natural questions.
Crowdsourcing data is challenging; adequate instruction must be quickly provided to crowdworkers. Workers want to maximize their earnings and will skip tasks with long, complicated directions. This difficulty is compounded by the fact that crowdworkers typically do not possess domain-specific knowledge.

In our case, we look for questions that meet certain criteria, such as: the question cannot give away the modality, the question cannot require specialized outside knowledge,
etc. If we just listed these stipulations, workers would skip the task due to its apparent complexity. Instead, we adopt a similar approach to \citet{VQA}, by framing our instructions as an adversarial game, in which the worker attempts to create a question that will fool a QA robot.\footnote{Example available in appendix.} We tell the crowdworkers that we have a robot capable of performing the task, manymodal QA over three modalities. Since we want the task to be difficult, we ask the workers to try to fool the robot to the best of their ability.
A critical component of our task is question ambiguity, therefore we describe what an ambiguous question is and again ask the workers to fool the robot by making the questions as ambiguous as possible. We find that this adversarial approach results in many of our criteria being met, without explicitly listing them.\footnote{During collection, we noticed that some image questions were simplistic and workers had difficulty creating ambiguous questions. For these reasons, we decided to provide captions with each of the images during collection and include them in our challenge. The intention of the captions is that the question still requires the image to be answered, but the caption provides additional context to make the questions more complex and ambiguous. We manually went through all image examples, filtering to ensure that image questions were not exclusively answerable using only the caption.} This makes the task less intimidating for the crowdworkers because the instructions are concise and easy to understand.

We validate our data with another MTurk task, where the crowdworkers are asked to produce a single word answer to a given question.
Each example is validated by 3 workers and is considered valid if at least 2 agree on the answer. We obtain a \textbf{92.58\% inter-annotator agreement} score, indicating that our data is high quality and answerable by humans.

We paid workers \$0.06 for each created image question-answer pair and estimated the average completion time to be 15-30 seconds. For text and table pairs, we paid \$0.05 with a 10-25 second estimated completion time. We pay more for image questions because we found that creating image questions that appear to be text/table questions is more time-consuming than vice-versa. This is because image questions tend to be visually focused and this sort of information is frequently not present in paragraphs or tables. To complete any of our HITs, workers had either 15 or 30 minutes for completion. Additionally, all tasks had the following worker qualifications: 1,000 previously completed HITs, 96\% HIT acceptance rate, and be a native-english speaker.

\section{Data Analysis}

\begin{figure}[t]
    \centering
    \includegraphics[width=0.4\textwidth]{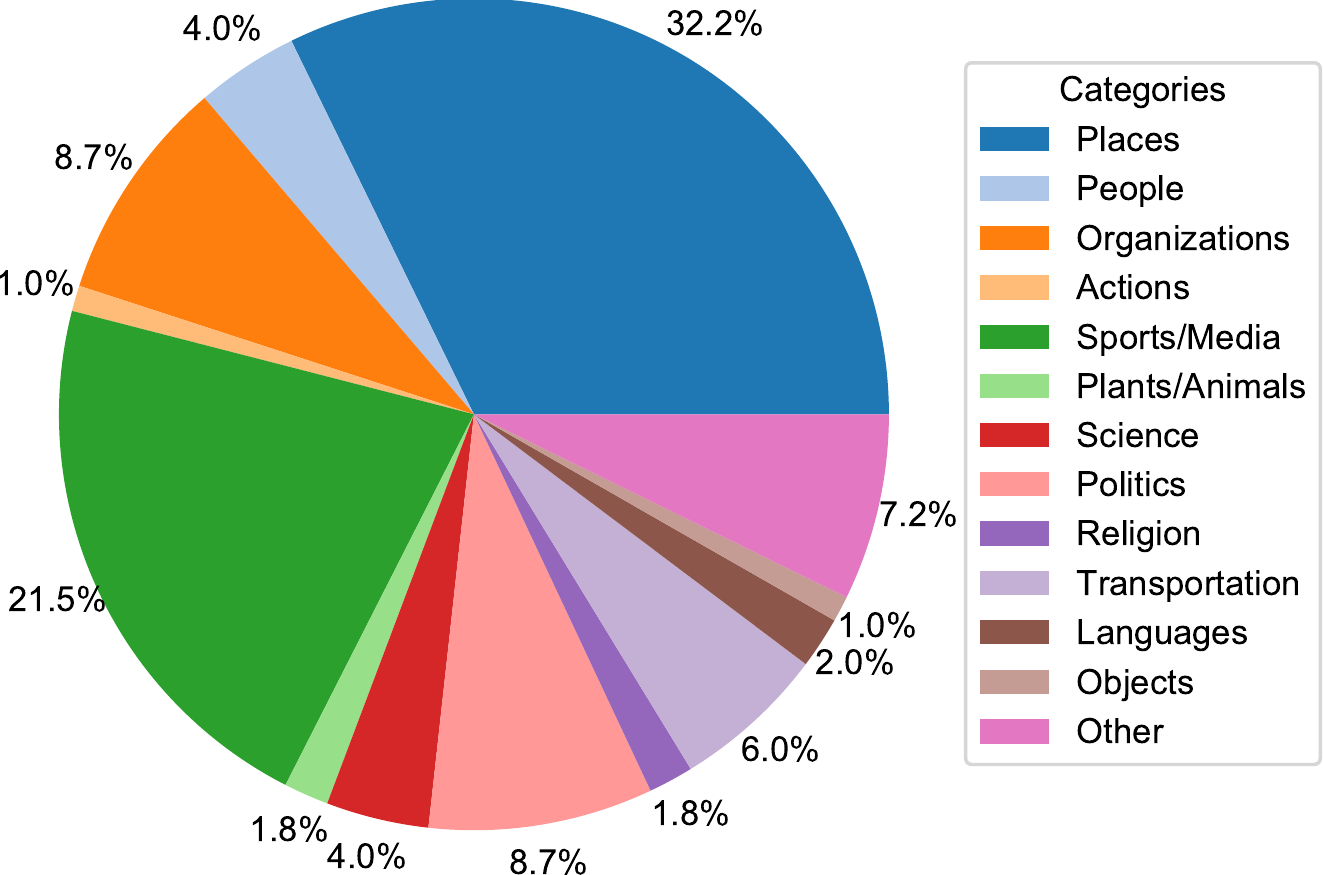}
    \caption{Distribution of Wikipedia pages across different categories, showing the breadth of our challenge.}
    \label{fig:breadth}
\end{figure}

\begin{figure*}[t]
    \centering
    \includegraphics[width=0.78\textwidth]{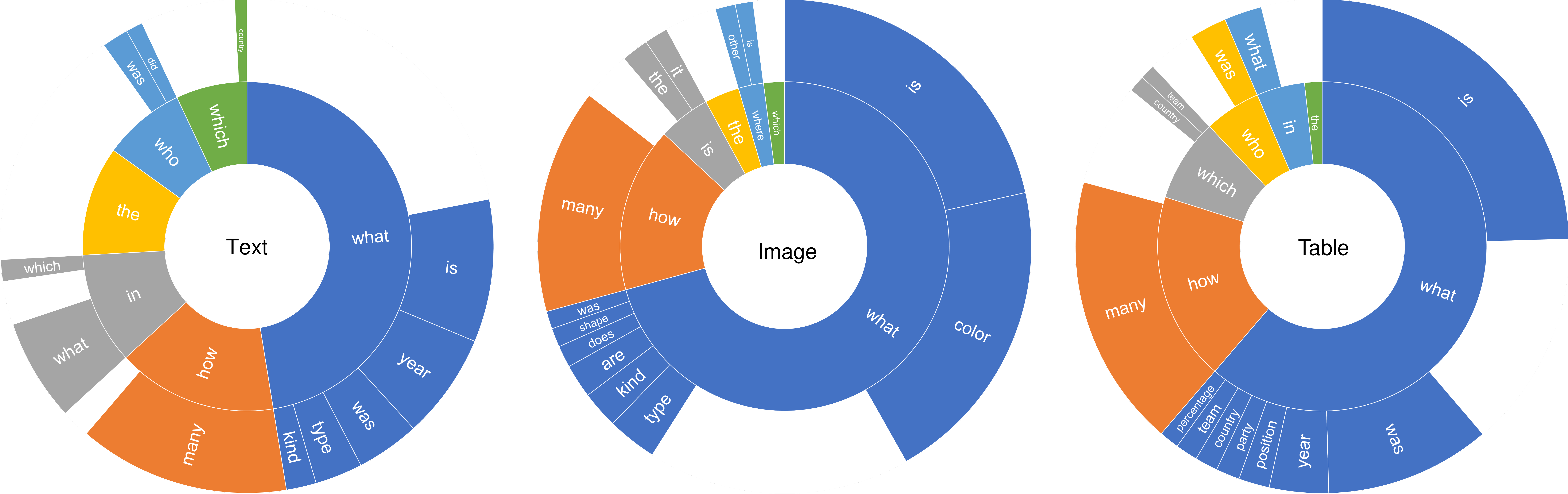}
    \caption{Distribution of questions for 6 most common first words and subsequent second words for each modality.}
    \label{fig:sunburst}
\end{figure*}

\modelname{} contains 10,190 questions: 2,873 image, 3,789 text, and 3,528 table; a 20/30/50\% fine-tuning/dev/test split. We select these splits to maximize the size of our dev and test sets, while using our challenge set only for fine-tuning. In all experiments, we tune our models on dev and only use test for final evaluation. Additionally, all examples in the paper are from train or dev. We publicly release the fine-tuning and the dev sets and keep the test set to evaluate submissions to our upcoming leaderboard.

\subsection{Question Properties}

\noindent\textbf{Sample Questions}: Figure \ref{tab:sample_questions} contains sample questions from our challenge set. Most of the questions are ambiguous with respect to the modality that was used to generate the question. In many cases, the set of potential modalities can be narrowed down to two. For instance, there are not any indications that table question 1 is indeed a table question, one can easily imagine a text paragraph that would contain the answer. However, it is difficult to imagine an image that could answer the question. Therefore, the model would have to search both the table and the page's text to find the answer.

\noindent\textbf{Knowledge Breadth}: Wikipedia contains 6 million articles, spanning many domains. As a result, our data covers many topics and models must handle this diverse information. To quantify this diversity, we categorize pages as one of 13 categories, seen in Figure \ref{fig:breadth}. Places and sports/media are our largest categories, at 53.7\%. This is likely due to these pages frequently containing tables, the least common modality, and thus frequently contain all 3, providing high-quality examples. Our data also contains complex examples, with 4.0\% and 8.7\% science and political pages, respectively.

The breadth of our dataset extends beyond just these categories. While the topic of a page corresponds to them, many pages contain a broad spectrum of knowledge in and of themselves. For instance, places is mostly composed of pages about countries, states, and cities. Each of these pages have many different sections about the history of the location, demographics, weather, landmarks, etc.

\begin{table}[t]
    \begin{small}
    \centering
    \begin{tabular}{|c|c||c|c|}
        \hline
         Data Type & Existing Datasets &  Our Dev & Our Test \\\hline
         \hline
         Images & 98.00\% & 75.58\%  & 75.68\% \\
         Text & 89.74\% & 78.54\%  & 77.72\% \\
         Tables & 90.37\% & 79.21\%  & 78.71\% \\ \hline
         Total & 92.65\% & 77.93\%  & 77.49\% \\\hline
    \end{tabular}
    \caption{Accuracy breakdown of the modality selection network. The first column is the accuracy on the unimodal datasets, and the rest are on our data.}
    \label{tab:msn}
    \end{small}
\end{table}

\noindent\textbf{Question Types}: We perform another analysis of the questions by considering the most common words in each modality; Figure~\ref{fig:sunburst} shows sunburst plots of question types contained in our data. There is not a large discrepancy between modalities because the most common types across each are similar. For instance, ``what" is the most common type in all three modalities, ``how many" is also common across all three, etc. It is only at a more fine-grained level that differences become apparent. This balance indicates that question type is a poor feature for determining the modality.

\subsection{Answer Properties}
There are 5,032 unique answers in our data. To maintain a diverse set, we limit the number of yes/no answers in the dataset, which make up 2.41\%. For text and table questions, we did not stipulate whether the answer needed to appear in the context, however we find that in 6.33\% of text-based examples and in 9.38\% of table-based examples, the answer does. This makes both tasks open-ended, as opposed to span-based. In span-based QA, the answer span always appears in the context document, and the model needs to determine the span. Open-ended QA is a more difficult task due to the answers being open-ended and makes our task more challenging, while also allowing the text and table modalities to mix with the visual modality, which can only be open-ended.

\section{Baseline Models and Results} 
We present a baseline model that combines existing state-of-the-art QA architectures for each of our three modalities, taking advantage of prior work in unimodal QA. This model not only serves as a baseline but also as a tool for analyzing the ambiguity of our questions. We hope that our challenge will encourage several end-to-end models in future work. First, we introduce our modality selection (or disambiguation) network that classifies each question as one of three modalities: text, image, or table.
We then use the modality selection network with existing QA models to create a modality selection and question-answering network (our \modelname{} model), that first runs the modality selector and then runs a pre-trained unimodal QA model corresponding to the modality that was selected. Without the modality selection network more rudimentary methods, for selecting which unimodal model to run would need to be used (such as the voting baseline presented in Section \ref{sec:modality_selection_results}.)
We also explore preliminary approaches to re-purposing and combining existing datasets in a manymodal setting.

\subsection{Modality Selection Network} \label{sec:modality_selection_desc}
We use our modality selection network to predict which modality the answer is found in. We frame this as a multi-class classification task where the network receives a question as input and predicts one of three classes corresponding to our 3 modalities.
The modality selection network receives the question as input and embeds it using pre-trained ELMo \citep{elmo}. The embeddings are fed through an LSTM and the resulting vector is passed through two feed-forward layers with a final softmax activation.
The model is trained exclusively on our fine-tuning set, tuned on our dev set, and test is only used for evaluation.
To evaluate the ambiguity of our questions, we compare the results on our data to results on data composed of 3 unimodal datasets: SQuAD v1.1 \citep{Rajpurkar2016SQuAD10} for text, VQA v2 \citep{balanced_vqa_v2} for image, and WikiTableQuestions \citep{wikitables} for table questions. For each dataset, we extract the questions and create labels that indicate which modality/dataset each belongs to. We make a training set with 1,800 examples, a dev set with 5,850, and a test set with 7,350. Since the training size is slightly smaller than our challenge set, if the model obtains higher accuracy on this data compared to ours, it suggests that our questions are more real-world ambiguous (see results below in Sec.~\ref{sec:modality_selection_results}).

\begin{figure}[t]
    \centering
    \includegraphics[width=0.4\textwidth]{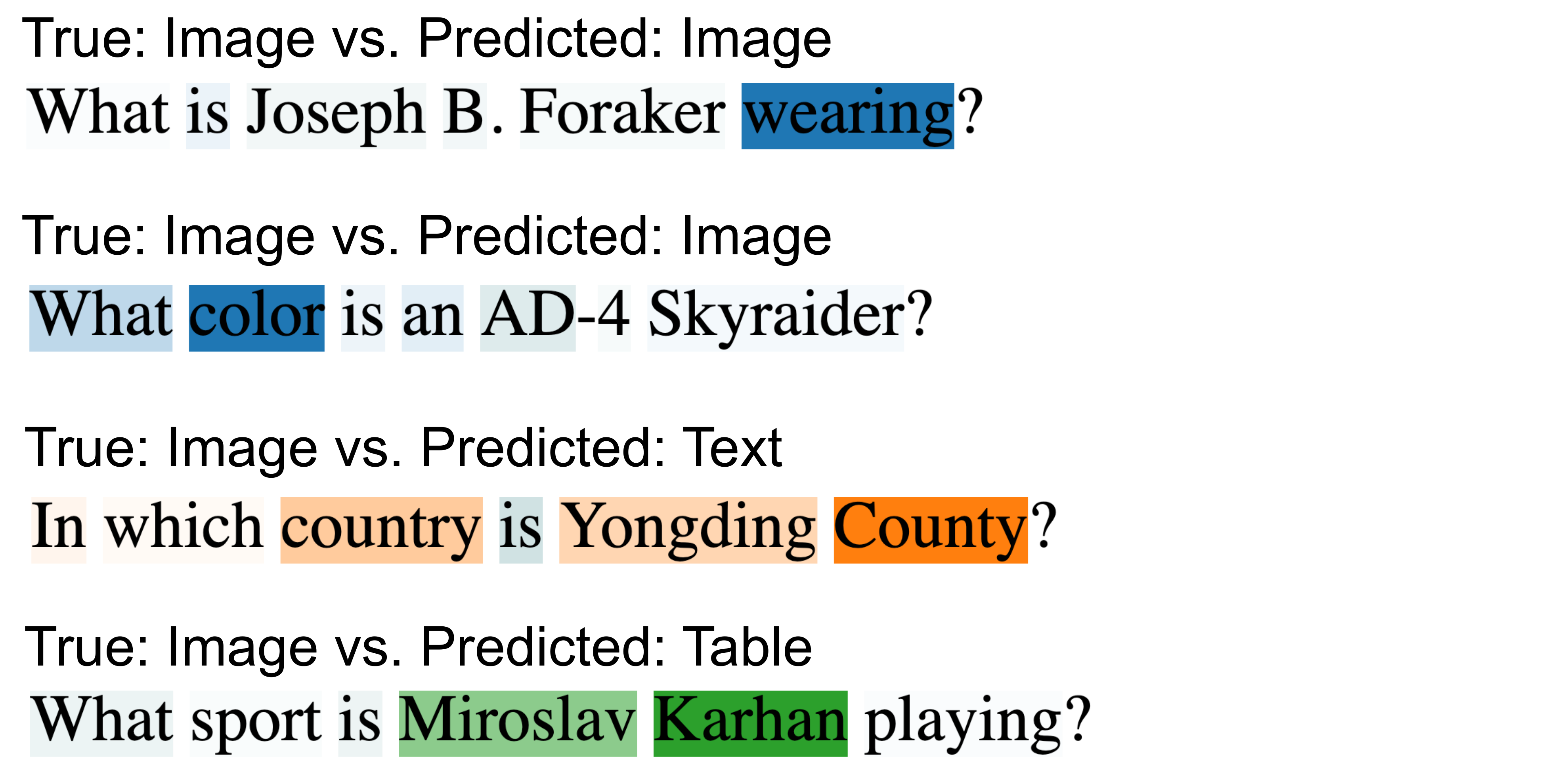}
    \caption{LIME heatmaps of image questions. Dark blue indicates a more positive contribution towards image, green indicates a more positive contribution towards tables, and orange indicates a more positive contribution towards text. }
    \label{fig:lime}
\end{figure}

\subsection{Modality Selection Network Results} \label{sec:modality_selection_results}

Table \ref{tab:msn} contains the accuracies of the modality selection network. Our model gets 77.49\% total accuracy on our data.
We compare the model to a most common baseline, where we always select the most common answer in the fine-tune data, achieving 37.14\%.
On the data composed of 3 unimodal datasets, our model gets 92.65\%. The model performs significantly worse on \modelname{} than on the unimodal data, supporting our hypothesis that the questions in our challenge set are more ambiguous.

\noindent\textbf{LIME Analysis}: To visualize the modality selection network, we perform a LIME (Locally Interpretable Model-Agnostic Explanations) analysis \citep{lime} (results in Figure \ref{fig:lime}). The top two sentences contain examples that were correctly classified as an image question with high confidence. From the resulting heatmap, it can be seen that the words ``color", and ``wearing" (visually oriented words) are important for classifying image questions, whereas ``Joseph B. Foraker" and ``Skyraider", proper nouns, are not. The bottom two sentences are also image questions, but were misclassified as text and table, respectively. The heatmaps show that ``Yongding County" and ``Miroslav Karhan" contribute to the misclassification, as proper nouns are more common in text/table questions.

There are cases where the question's classification seems intuitively correct, yet is actually incorrect. Consider the question ``What color stone is the exterior of the church made from?''. This question was classified as an image question due to it inquiring about a visual feature of the church, however, it is actually a text question. The only way that a model can classify questions like this correctly, is by reasoning over the input modalities and the question. However, as explained in the following section, this is a non-trivial problem that our task encapsulates to encourage the community to pursue this as future work. An additional analysis in the form of a confusion matrix is in the appendix.

\begin{figure}[t]
    \centering
    \includegraphics[width=0.32\textwidth]{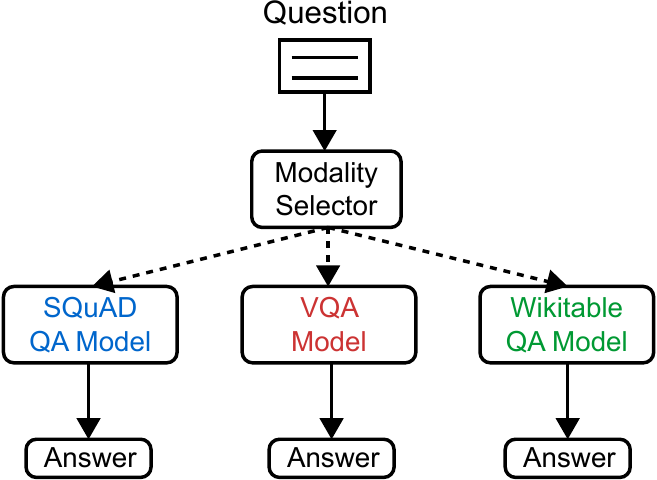}

    \caption{Our baseline network architecture for modality selection and question answering.}
    \label{fig:predictor_architecture}
\end{figure}

\subsection{Modality Selection and Question-Answering Network} \label{sec:msqa_desc}
In this model, we have 4 separate sub-models: a modality selection network and one unimodal QA model for each modality. The modality selection network predicts which modality the answer will be found in, as described in Section \ref{sec:modality_selection_desc}. Then, based upon the prediction, the appropriate context (text, table, or image) and question are fed to the corresponding unimodal model (Figure \ref{fig:predictor_architecture}). Note this is not a full end-to-end model. This makes our architecture more robust because any unimodal model can be used. We have  structured our challenge set in a way that encourages the community to develop flexible end-to-end models that can generalize across many modalities by properly combining them in a low-resource setting, improving upon existing transfer learning techniques. Since there are no existing large pre-training datasets available that contain all three modalities together, one may utilize separate datasets for each modality (as we do in this work), but if the model is end-to-end, then more advanced training regimes will be required to handle missing modalities. Thus, this presents an interesting challenge for future work, where training examples consist of only a single modality, yet the model must learn to handle multiple inputs because our test examples are manymodal.

All networks are trained separately. For each modality, a model is trained on an existing unimodal dataset. Each architecture achieves state-of-the-art, or near state-of-the-art, in its respective domain. The modality selection network is trained as described in Section \ref{sec:modality_selection_desc}. At test time, the question is run through this network, a modality is predicted, and the corresponding model generates an answer. All models are fine-tuned on our fine-tuning set, tuned on our dev set, and our test set is used only for evaluation. Note that fine-tuning data is not passed through the modality selector when fine-tuning the unimodal models; ground-truth labels are used.

\noindent{\bf{Text-based QA Model}} \label{sec:text_model}
For our text-based questions, we use a state-of-the-art BERT QA model with a pointer-generator mechanism \citep{Devlin2018BERTPO,pointergen}. We train the model on a version of SQuAD v1.1 that we modified to better match our data \citep{Rajpurkar2016SQuAD10}. We select SQuAD v1.1 because it is also Wikipedia-based and has been established as a high-quality dataset. We did not use SQuAD v2.0 because we do not have unanswerable questions in our challenge set.
We define the probability of producing a given word as follows:
\begin{equation}
    P(w) = p_{gen} P_{ans}(w) + (1 - p_{gen}) P_{sel}(w)
\end{equation}
where $p_{gen}$ is the probability of generating a word from our candidate set versus selecting one, $P_{ans}$ is the probability of generating a given answer from our candidate set, and $P_{sel}$ is the probability of selecting a given word from the document (see appendix for more information).
Since SQuAD does not contain yes/no questions, we use simple heuristics to convert a small percentage of questions to yes/no. Then, we discard any answers that are longer than a single word, since every answer in our challenge set consists of a single word, leaving 25,504 training examples.

\noindent{\bf{Table-based QA Model}}
To process our tables, we use Stanford's SEMPRE framework, trained on WikiTableQuestions \citep{wikitables} and fine-tuned on our data, which is a state-of-the-art approach based on semantic parsing.
We also experiment with some alternative approaches to table QA, namely, we use the text BERT model, treating our tables like text and pre-training the model on additional tables that we collect from Wikipedia, but this approach does not perform as well and is a good venue for future work.

\noindent{\bf{Image-based QA Model}}
Our imageQA model uses the bottom-up attention architecture \citep{bottom-up} and is trained using VQA v2~\citep{balanced_vqa_v2}. We make a minimal modification to the model in the form of an additional LSTM with GloVe \citep{glove} (just like the question) to process the caption. The output of the LSTM is concatenated with the joint representation of the image and question.
We force the candidate answers to contain every answer in our fine-tuning set, plus any answer that appears more than 20 times in VQA v2. This ensures that the set of candidate answers does not get too large but still contains ours.
To get captions during training, we use each image's corresponding example in VisDial \citep{visdial} because it is also based on COCO \citep{COCO} and includes one of the COCO captions for each example.

\subsection{Modality Selection and Question Answering Network Results} \label{sec:5.4}

The results for our modality selection and answering network are in Table \ref{tab:mpn}. The model obtains 39.73\% accuracy on our test data. We tune the model on our dev data and get 26.98\% on images, 49.87\% on text, and 39.13\% on tables, with a total accuracy of 39.70\% on the dev set. We also evaluate an oracle model without the modality selector, where each question is fed to the appropriate model using its ground-truth modality, getting 46.51\% and 46.26\% for dev and test, respectively. We conduct human evaluation on our data, where a human is asked to answer the question given all modalities. Due to the open-ended nature of our task, we manually evaluate the responses. Humans achieve \textbf{91.58\%} accuracy, indicating that our data is high-quality and that current state-of-the-art techniques are insufficient, leaving a lot of room for community improvements. We also compare this model to three baselines. The first is a voting baseline where each unimodal model is run for all available modalities in each example, then if 2 or more models guess the same answer, that answer is returned; otherwise, an answer is returned randomly from the three models. This approach gets 21.12\% on our test data, demonstrating the importance of the modality selector. We also have two other simple baselines: a `most-common' baseline, which selects the most common answer from our fine-tuning data, for every example in the test data, obtaining 2.41\%; and a nearest neighbor baseline, which answers a given question in the test set with the answer from the closest matching embedding from the universal sentence encoder (USE) \citep{universal_encoder} in the fine-tuning set, achieving 5.61\% accuracy. These results (in conjunction with the high human performance) confirm that \modelname{} is a challenging test bed for future work.

\begin{table}[t]
    \centering
    \begin{small}
    \begin{tabular}{|c|c|c|c|c|}
        \hline
         Model  & Images & 
         Text & 
         Tables &
         Total\\
         \hline
         \hline
         Most Common & 4.17\% & 1.58\% & 1.87\% & 2.41\% \\
         
         USE & 11.40\% & 3.70\% & 2.94\% & 5.61\% \\
         
         Voting & 15.50\% & 23.71\% & 22.93\% & 21.12\% \\\hline

         Our Model & 27.17\% & 48.63\% & 40.43\% & 39.73\%\\

         Oracle & 29.05\% & 59.35\% & 46.26\% & 46.26\%\\\hline

        Human & 94.00\% & 92.00\% & 89.61\% & 91.58\%
         \\\hline

    \end{tabular}
    \caption{Accuracy (EM) breakdown of the modality selection and answering  network on our test data. }
    \label{tab:mpn}
    \end{small}
\end{table}

\subsection{Recent Results with RoBERTa and LXMERT}
Here, we present new experiments that utilize very recent modeling techniques, improving the performance of our baselines, but still leaving plenty of room for improvement.
For the modality selector, we have upgraded ELMo to RoBERTa \citep{Liu2019RoBERTaAR}, a large-scale transformer-based model. This change increases the performance (in Table \ref{tab:msn}) on our dev data to 80.92\% (from 77.93\%) and on our test data to 81.35\% (from 77.49\%), while the performance on existing datasets is increased to 95.44\% (from 92.65\%). These increased numbers indicate that with improved language modeling, questions can be more successfully disambiguated; however, our dataset is still substantially more ambiguous, and there still remain questions - such as the example discussed in Section \ref{sec:modality_selection_results} - that cannot be disambiguated without considering the modality contexts.
In our textQA model, we replaced BERT \citep{Devlin2018BERTPO} again with RoBERTa \citep{Liu2019RoBERTaAR}, but it does not have a large impact on performance. We also the latest LXMERT~\citep{tan2019lxmert} method to our imageQA model; this yields a significant improvement, increasing the score in Table \ref{tab:mpn} to 37.11\% (from 27.17\%) and increasing the oracle score to 40.79\% (from 29.05\%), demonstrating the usefulness of recent large-scale multimodal pretraining but still maintaining a large gap to fill by future work.

\section{Conclusion} \label{conclusion}
We presented a new challenge to the community, hoping to promote research in manymodal QA. We structured our challenge in a way that encourages research in other, more general areas, such as transfer learning and end-to-end modality disambiguation + multimodal QA. We hope that this challenge will serve as a test bed for further work and that our model will inspire directions of subsequent research. We plan to continue our work by collecting data that exhibits a stronger form of multimodality, where the question can only be answered after combining multiple modalities, and by adding new modalities, such as video and audio.

\section{Acknowledgements} \label{Acknowledgements}
We  thank  the  reviewers  for  their  helpful  comments. This work   was   supported   by  DARPA MCS Grant \#N66001-19-2-4031  ARO-YIP  Award  \#W911NF-18-1-0336, an NSF PhD Fellowship, and faculty awards from Google and Facebook. The  opinions are of the authors, not the  funding agency.

\fontsize{9.0pt}{10.0pt} \selectfont
\bibliographystyle{aaai}
\bibliography{aaai2020}

\begin{thebibliography}{}

\bibitem[\protect\citeauthoryear{Anderson \bgroup et al\mbox.\egroup
  }{2018}]{bottom-up}
Anderson, P.; He, X.; Buehler, C.; Teney, D.; Johnson, M.; Gould, S.; and
  Zhang, L.
\newblock 2018.
\newblock Bottom-up and top-down attention for image captioning and visual
  question answering.
\newblock In {\em CVPR}.

\bibitem[\protect\citeauthoryear{Antol \bgroup et al\mbox.\egroup }{2015}]{VQA}
Antol, S.; Agrawal, A.; Lu, J.; Mitchell, M.; Batra, D.; Zitnick, C.~L.; and
  Parikh, D.
\newblock 2015.
\newblock {VQA}: {V}isual {Q}uestion {A}nswering.
\newblock In {\em ICCV}.

\bibitem[\protect\citeauthoryear{Cer \bgroup et al\mbox.\egroup
  }{2018}]{universal_encoder}
Cer, D.; Yang, Y.; Kong, S.; Hua, N.; Limtiaco, N.; John, R.~S.; Constant, N.;
  Guajardo{-}Cespedes, M.; Yuan, S.; Tar, C.; Sung, Y.; Strope, B.; and
  Kurzweil, R.
\newblock 2018.
\newblock Universal sentence encoder.
\newblock {\em CoRR} abs/1803.11175.

\bibitem[\protect\citeauthoryear{Cho \bgroup et al\mbox.\egroup
  }{2018}]{cho2018adversarial}
Cho, M.; Amplayo, R.~K.; Hwang, S.-w.; and Park, J.
\newblock 2018.
\newblock Adversarial tableqa: Attention supervision for question answering on
  tables.
\newblock {\em ACML}.

\bibitem[\protect\citeauthoryear{Clark and Gardner}{2018}]{Clark2018SimpleAE}
Clark, C., and Gardner, M.
\newblock 2018.
\newblock Simple and effective multi-paragraph reading comprehension.
\newblock In {\em ACL}.

\bibitem[\protect\citeauthoryear{Das \bgroup et al\mbox.\egroup
  }{2017}]{visdial}
Das, A.; Kottur, S.; Gupta, K.; Singh, A.; Yadav, D.; Moura, J.~M.; Parikh, D.;
  and Batra, D.
\newblock 2017.
\newblock {V}isual {D}ialog.
\newblock In {\em CVPR}.

\bibitem[\protect\citeauthoryear{Devlin \bgroup et al\mbox.\egroup
  }{2018}]{Devlin2018BERTPO}
Devlin, J.; Chang, M.-W.; Lee, K.; and Toutanova, K.
\newblock 2018.
\newblock Bert: Pre-training of deep bidirectional transformers for language
  understanding.
\newblock In {\em NAACL-HLT}.

\bibitem[\protect\citeauthoryear{Dunn \bgroup et al\mbox.\egroup
  }{2017}]{searchqa}
Dunn, M.; Sagun, L.; Higgins, M.; Guney, V.~U.; Cirik, V.; and Cho, K.
\newblock 2017.
\newblock Searchqa: A new q\&a dataset augmented with context from a search
  engine.
\newblock {\em CoRR} abs/1704.05179.

\bibitem[\protect\citeauthoryear{Goyal \bgroup et al\mbox.\egroup
  }{2017}]{balanced_vqa_v2}
Goyal, Y.; Khot, T.; Summers{-}Stay, D.; Batra, D.; and Parikh, D.
\newblock 2017.
\newblock Making the {V} in {VQA} matter: Elevating the role of image
  understanding in {V}isual {Q}uestion {A}nswering.
\newblock In {\em CVPR}.

\bibitem[\protect\citeauthoryear{Hermann \bgroup et al\mbox.\egroup
  }{2015}]{cnndailymail}
Hermann, K.~M.; Kocisk{\'y}, T.; Grefenstette, E.; Espeholt, L.; Kay, W.;
  Suleyman, M.; and Blunsom, P.
\newblock 2015.
\newblock Teaching machines to read and comprehend.
\newblock In {\em NeurIPS}.

\bibitem[\protect\citeauthoryear{Jauhar, Turney, and
  Hovy}{2016}]{Jauhar2016TablesAS}
Jauhar, S.~K.; Turney, P.~D.; and Hovy, E.~H.
\newblock 2016.
\newblock Tables as semi-structured knowledge for question answering.
\newblock In {\em ACL}.

\bibitem[\protect\citeauthoryear{Johnson \bgroup et al\mbox.\egroup
  }{2017}]{clevr}
Johnson, J.; Hariharan, B.; van~der Maaten, L.; Fei-Fei, L.; Zitnick, C.~L.;
  and Girshick, R.~B.
\newblock 2017.
\newblock Clevr: A diagnostic dataset for compositional language and elementary
  visual reasoning.
\newblock In {\em CVPR}.

\bibitem[\protect\citeauthoryear{Kafle \bgroup et al\mbox.\egroup
  }{2018}]{Kafle2018DVQAUD}
Kafle, K.; Cohen, S.; Price, B.~L.; and Kanan, C.
\newblock 2018.
\newblock Dvqa: Understanding data visualizations via question answering.
\newblock In {\em CVPR}.

\bibitem[\protect\citeauthoryear{Kembhavi \bgroup et al\mbox.\egroup
  }{2017}]{textbookqa}
Kembhavi, A.; Seo, M.; Schwenk, D.; Choi, J.; Farhadi, A.; and Hajishirzi, H.
\newblock 2017.
\newblock Are you smarter than a sixth grader? textbook question answering for
  multimodal machine comprehension.
\newblock In {\em CVPR}.

\bibitem[\protect\citeauthoryear{Kocisky \bgroup et al\mbox.\egroup
  }{2018}]{narrativeqa}
Kocisky, T.; Schwarz, J.; Blunsom, P.; Dyer, C.; Hermann, K.~M.; Melis, G.; and
  Grefenstette, E.
\newblock 2018.
\newblock The narrativeqa reading comprehension challenge.
\newblock {\em TACL}.

\bibitem[\protect\citeauthoryear{Lai \bgroup et al\mbox.\egroup
  }{2017}]{raceqa}
Lai, G.; Xie, Q.; Liu, H.; Yang, Y.; and Hovy, E.~H.
\newblock 2017.
\newblock Race: Large-scale reading comprehension dataset from examinations.
\newblock In {\em EMNLP}.

\bibitem[\protect\citeauthoryear{Lei \bgroup et al\mbox.\egroup
  }{2018}]{lei2018tvqa}
Lei, J.; Yu, L.; Bansal, M.; and Berg, T.~L.
\newblock 2018.
\newblock Tvqa: Localized, compositional video question answering.
\newblock In {\em EMNLP}.

\bibitem[\protect\citeauthoryear{Lin \bgroup et al\mbox.\egroup }{2014}]{COCO}
Lin, T.-Y.; Maire, M.; Belongie, S.~J.; Bourdev, L.~D.; Girshick, R.~B.; Hays,
  J.; Perona, P.; Ramanan, D.; Doll{\'a}r, P.; and Zitnick, C.~L.
\newblock 2014.
\newblock Microsoft coco: Common objects in context.
\newblock In {\em ECCV}.

\bibitem[\protect\citeauthoryear{Liu \bgroup et al\mbox.\egroup
  }{2019}]{Liu2019RoBERTaAR}
Liu, Y.; Ott, M.; Goyal, N.; Du, J.; Joshi, M.; Chen, D.; Levy, O.; Lewis, M.;
  Zettlemoyer, L.~S.; and Stoyanov, V.
\newblock 2019.
\newblock Roberta: A robustly optimized bert pretraining approach.
\newblock {\em ArXiv} abs/1907.11692.

\bibitem[\protect\citeauthoryear{Logan, Humeau, and
  Singh}{2017}]{RobertLLogan2017MultimodalAE}
Logan, I. R.~L.; Humeau, S.; and Singh, S.
\newblock 2017.
\newblock Multimodal attribute extraction.
\newblock In {\em CoRR}, volume abs/1711.11118.

\bibitem[\protect\citeauthoryear{Maharaj \bgroup et al\mbox.\egroup
  }{2017}]{Maharaj2017ADA}
Maharaj, T.; Ballas, N.; Courville, A.~C.; and Pal, C.~J.
\newblock 2017.
\newblock A dataset and exploration of models for understanding video data
  through fill-in-the-blank question-answering.
\newblock In {\em CVPR}.

\bibitem[\protect\citeauthoryear{Nguyen \bgroup et al\mbox.\egroup
  }{2016}]{msmarco}
Nguyen, T.; Rosenberg, M.; Song, X.; Gao, J.; Tiwary, S.; Majumder, R.; and
  Deng, L.
\newblock 2016.
\newblock Ms marco: A human generated machine reading comprehension dataset.
\newblock In {\em NeurIPS}.

\bibitem[\protect\citeauthoryear{Pasupat and Liang}{2015}]{wikitables}
Pasupat, P., and Liang, P.
\newblock 2015.
\newblock Compositional semantic parsing on semi-structured tables.
\newblock In {\em ACL}.

\bibitem[\protect\citeauthoryear{Pennington, Socher, and Manning}{2014}]{glove}
Pennington, J.; Socher, R.; and Manning, C.~D.
\newblock 2014.
\newblock Glove: Global vectors for word representation.
\newblock In {\em EMNLP}.

\bibitem[\protect\citeauthoryear{Peters \bgroup et al\mbox.\egroup
  }{2018}]{elmo}
Peters, M.~E.; Neumann, M.; Iyyer, M.; Gardner, M.; Clark, C.; Lee, K.; and
  Zettlemoyer, L.
\newblock 2018.
\newblock Deep contextualized word representations.
\newblock In {\em NAACL}.

\bibitem[\protect\citeauthoryear{Rajpurkar \bgroup et al\mbox.\egroup
  }{2016}]{Rajpurkar2016SQuAD10}
Rajpurkar, P.; Zhang, J.; Lopyrev, K.; and Liang, P.
\newblock 2016.
\newblock Squad: 100, 000+ questions for machine comprehension of text.
\newblock In {\em EMNLP}.

\bibitem[\protect\citeauthoryear{Rajpurkar, Jia, and Liang}{2018}]{squad20}
Rajpurkar, P.; Jia, R.; and Liang, P.
\newblock 2018.
\newblock Know what you don't know: Unanswerable questions for squad.
\newblock In {\em ACL}.

\bibitem[\protect\citeauthoryear{Reddy, Chen, and Manning}{2018}]{coqa}
Reddy, S.; Chen, D.; and Manning, C.~D.
\newblock 2018.
\newblock Coqa: A conversational question answering challenge.
\newblock {\em TACL} 7:249--266.

\bibitem[\protect\citeauthoryear{Ribeiro, Singh, and Guestrin}{2016}]{lime}
Ribeiro, M.~T.; Singh, S.; and Guestrin, C.
\newblock 2016.
\newblock ``why should {I} trust you?": Explaining the predictions of any
  classifier.
\newblock In {\em KDD}.

\bibitem[\protect\citeauthoryear{Rohrbach \bgroup et al\mbox.\egroup
  }{2015}]{Rohrbach2015ADF}
Rohrbach, A.; Rohrbach, M.; Tandon, N.; and Schiele, B.
\newblock 2015.
\newblock A dataset for movie description.
\newblock In {\em CVPR}.

\bibitem[\protect\citeauthoryear{See, Liu, and Manning}{2017}]{pointergen}
See, A.; Liu, P.~J.; and Manning, C.~D.
\newblock 2017.
\newblock Get to the point: Summarization with pointer-generator networks.
\newblock In {\em ACL}.

\bibitem[\protect\citeauthoryear{Sun \bgroup et al\mbox.\egroup
  }{2016}]{SunCellSearch}
Sun, H.; Ma, H.; He, X.; Yih, W.-t.; Su, Y.; and Yan, X.
\newblock 2016.
\newblock Table cell search for question answering.
\newblock In {\em WWW}.

\bibitem[\protect\citeauthoryear{Tan and Bansal}{2019}]{tan2019lxmert}
Tan, H., and Bansal, M.
\newblock 2019.
\newblock Lxmert: Learning cross-modality encoder representations from
  transformers.
\newblock In {\em EMNLP}.

\bibitem[\protect\citeauthoryear{Welbl, Stenetorp, and Riedel}{2018}]{wikihop}
Welbl, J.; Stenetorp, P.; and Riedel, S.
\newblock 2018.
\newblock Constructing datasets for multi-hop reading comprehension across
  documents.
\newblock In {\em ACL}.

\bibitem[\protect\citeauthoryear{Yagcioglu \bgroup et al\mbox.\egroup
  }{2018}]{recipeqa}
Yagcioglu, S.; Erdem, A.; Erdem, E.; and Ikizler-Cinbis, N.
\newblock 2018.
\newblock Recipeqa: A challenge dataset for multimodal comprehension of cooking
  recipes.
\newblock In {\em EMNLP}.

\bibitem[\protect\citeauthoryear{Yang \bgroup et al\mbox.\egroup
  }{2018}]{yang2018hotpotqa}
Yang, Z.; Qi, P.; Zhang, S.; Bengio, Y.; Cohen, W.~W.; Salakhutdinov, R.; and
  Manning, C.~D.
\newblock 2018.
\newblock {HotpotQA}: A dataset for diverse, explainable multi-hop question
  answering.
\newblock In {\em EMNLP}.

\end{thebibliography}

\appendix
\section*{Appendices}

\section{Text-based BERT QA Model}
Before processing our data through our text-based model, we do some additional pre-processing where we use TF-IDF, similar to \cite{Clark2018SimpleAE}, to perform paragraph selection by selecting the paragraph with the closest TF-IDF vector to the question, shortening the context significantly.

The basis of the model is the same as the QA version of BERT described in~\cite{Devlin2018BERTPO}. However, due to the answers in our task being open-ended, we add a pointer-generator mechanism to the output layer. This allows the model to produce answers, such as ``yes" and ``no", that may not appear in the context document. More concretely, we define the probability of producing a given word as follows:
\begin{equation}
    P(w) = p_{gen} P_{ans}(w) + (1 - p_{gen}) P_{sel}(w)
\end{equation}
where $p_{gen}$ is the probability of generating a word from our candidate set versus selecting one, $P_{ans}$ is the probability of generating a given answer from our candidate set, and $P_{sel}$ is the probability of selecting a given word from the document.

$P_{sel}$ is produced by a linear layer on top of the pre-trained BERT model, acting as the pointer, producing a probability distribution over the words in the context document. To obtain $P_{ans}$, the output of the BERT model is fed through an LSTM to obtain a fixed-size representation, and then fed through a linear layer to produce a distribution over the set of candidate answers in the training set. $p_{gen}$ is produced by concatenating the fixed-size representation with the output of the pointer (prior to any softmax) and passing this through a linear layer.

\section{Question Lengths}
\begin{figure}[h]
    \centering
    \includegraphics[width=3in]{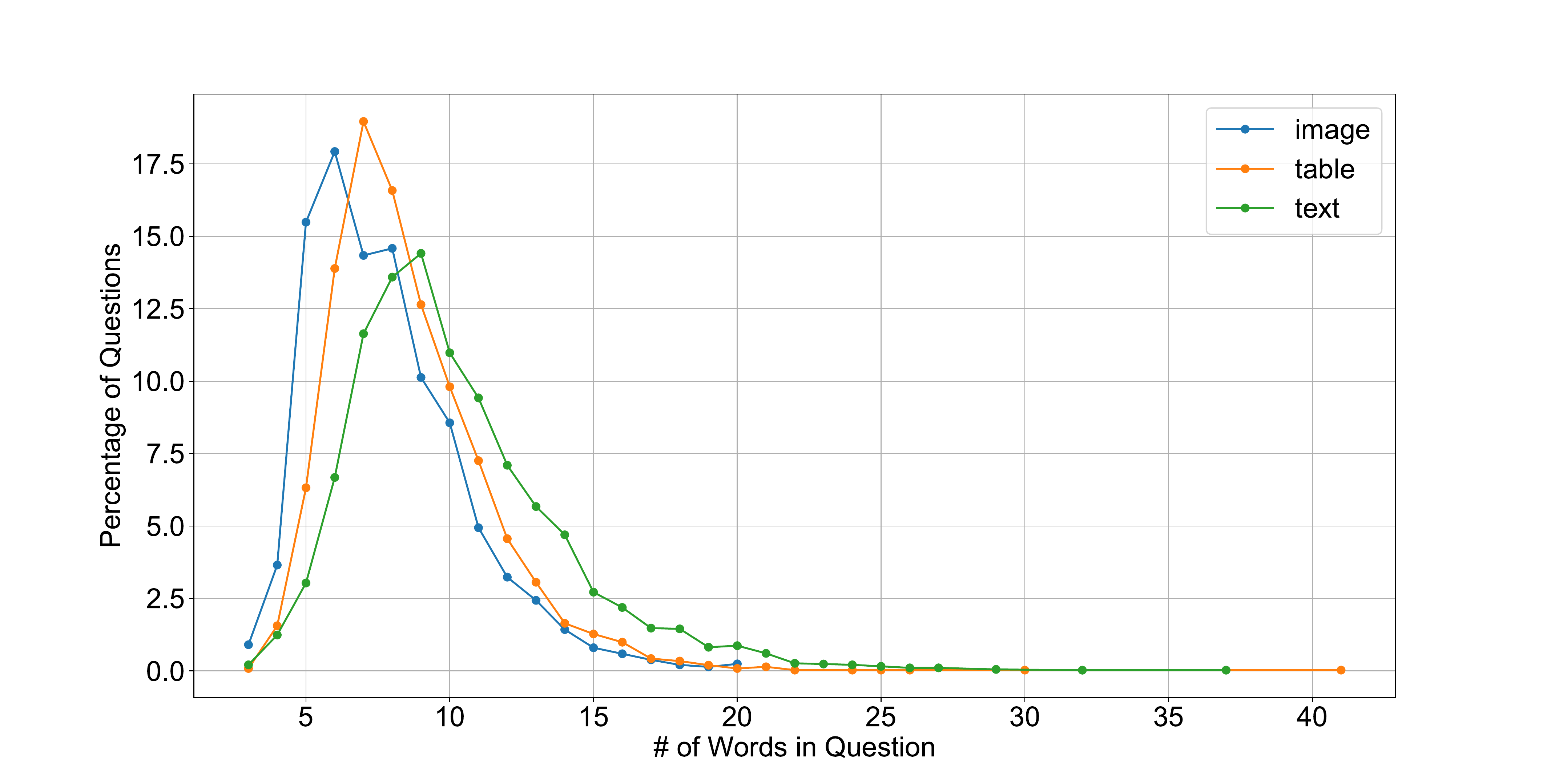}
    \caption{Percentage of questions with various sentence lengths for text, images, tables.}
    \label{fig:length}
\end{figure}
The number of words in the questions for each modality can be seen in Figure \ref{fig:length}. The distributions are relatively close to one another, with text questions being slightly longer, on average, than the other two modalities.

\begin{figure}[ht]
    \centering
    \includegraphics[width=3in]{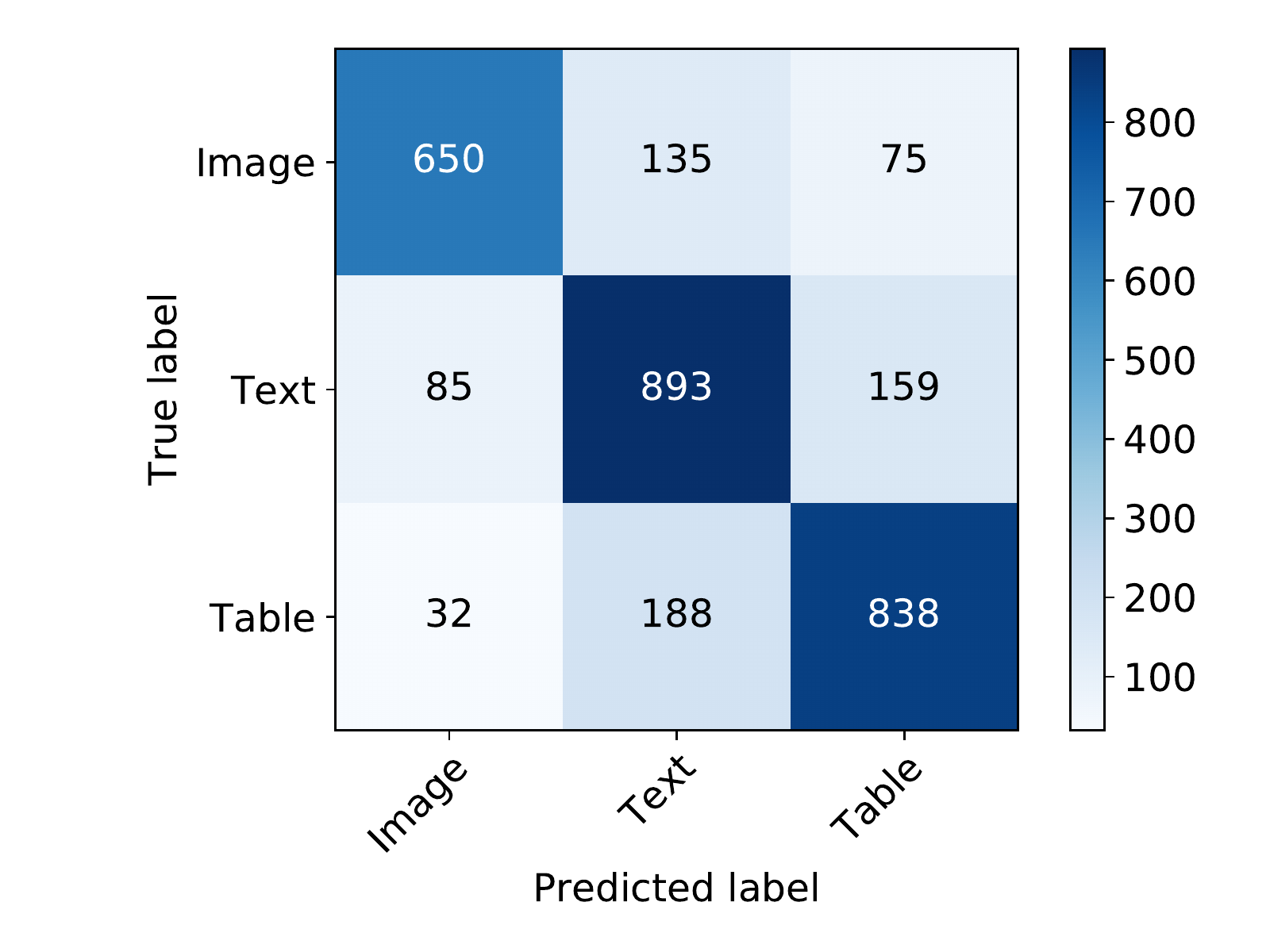}
    \caption{A confusion matrix from the dev data on our modality selector network.}
    \label{fig:conf_mat}
\end{figure}

\begin{figure*}[t]
    \centering
    \includegraphics[width=0.8\textwidth]{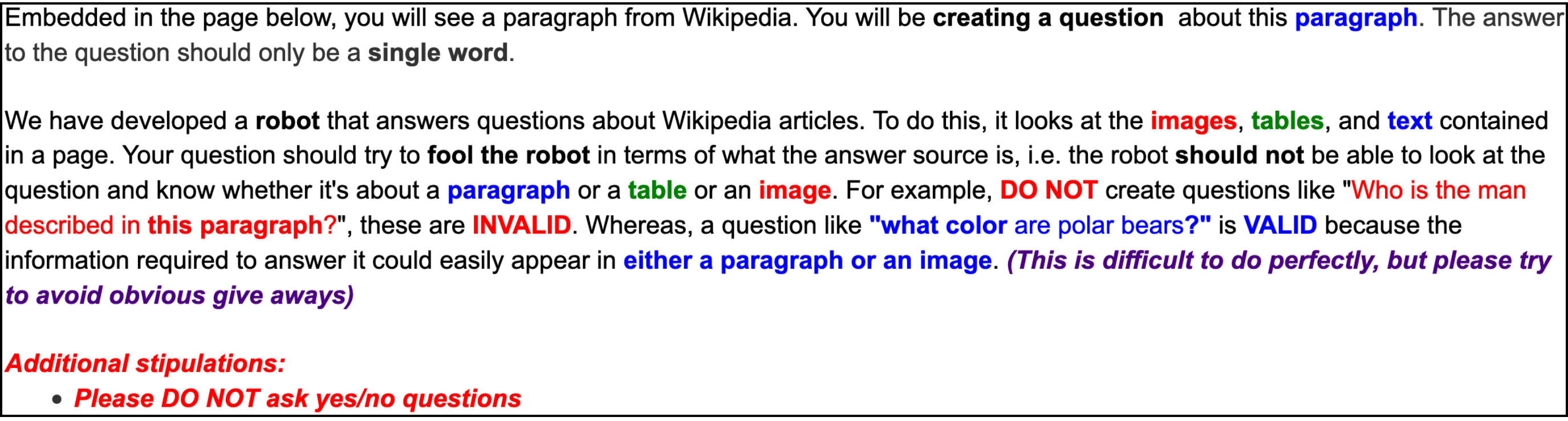}
    \caption{A sample of the instructions that we use for crowdsourcing.}
    \label{fig:crowd_instructions}
\end{figure*}

\begin{figure*}[t]
    \centering
    \includegraphics[width=0.8\textwidth]{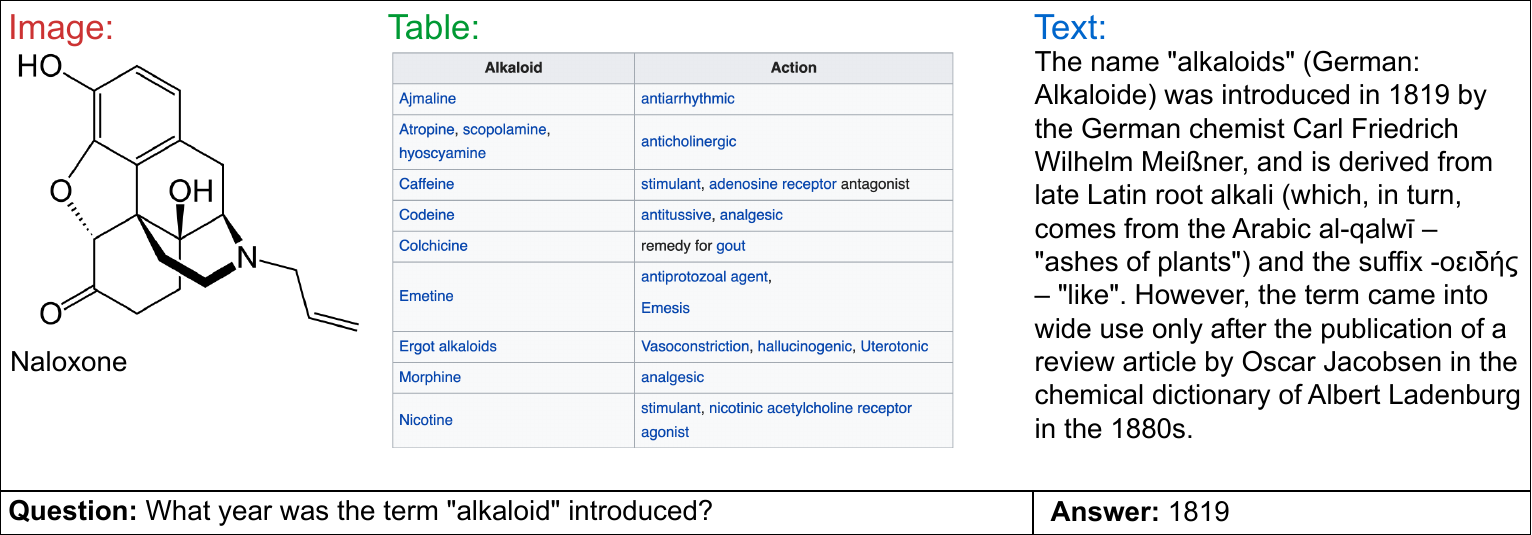}
    \caption{A condensed example of our task, using a sample that we collected from a Wikipedia page.}
    \label{fig:example1}
\end{figure*}

\begin{figure*}[h!]
    \centering
    \includegraphics[width=0.8\textwidth]{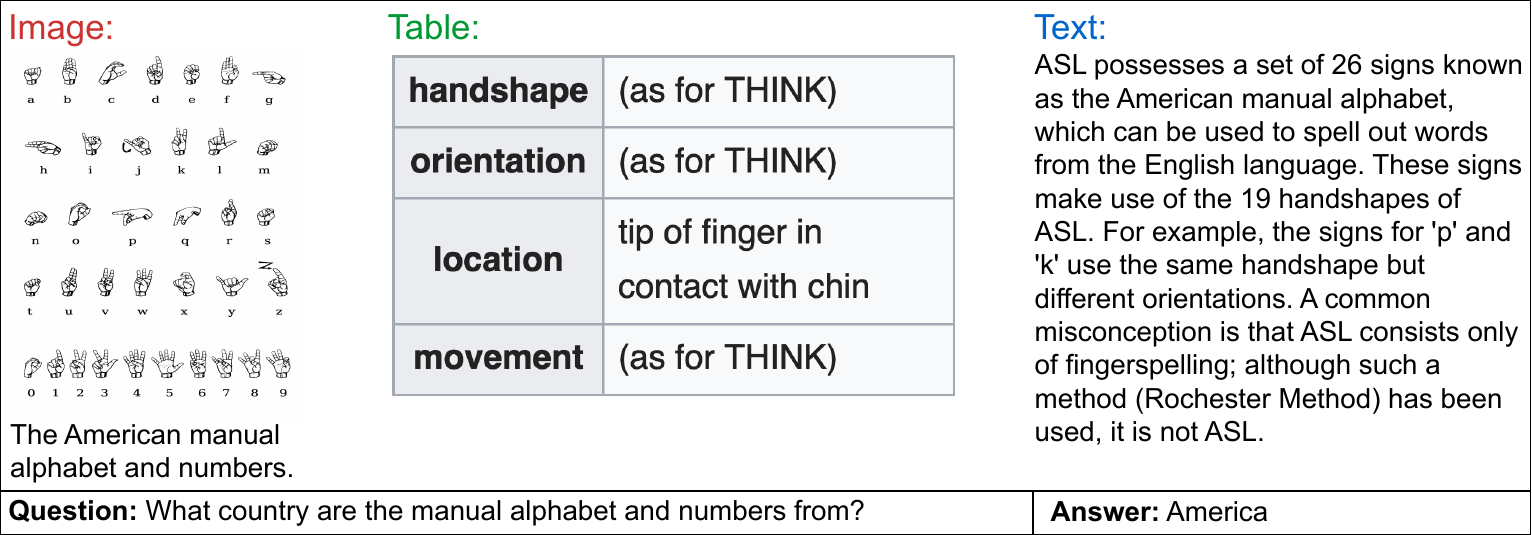}
    \caption{A condensed example of our task, using a sample that we collected from a Wikipedia page.}
    \label{fig:example2}
\end{figure*}

\section{Modality Selector Confusion Matrix}

Figure \ref{fig:conf_mat} contains a confusion matrix of our dev data passed through the modality selector. From this figure, it can be seen that text and table questions are frequently misclassified; which indicates that these modalities are the most similar. However, when image questions are misclassified, they tend to favor text. This is likely because text has the largest set of potential question types, as it is the most robust modality.

\section{Sample Mechanical Turk Instructions}

Figure \ref{fig:crowd_instructions} contains a sample of the instructions that we use to collect text QA examples. We use an adversarially-oriented approach where crowdworkers attempt to create questions that will fool a robot designed to complete our task. We emphasize the importance of question ambiguity in our instructions, as this is a critical part of our challenge set. Additionally, we try to keep the instructions as short as possible to increase the likelihood that they will be read completely.

\section{Data Samples}
Figures \ref{fig:example1} and \ref{fig:example2} show additional samples from our challenge set.
Figure \ref{fig:example1} demonstrates the breadth of our challenge set, with the question coming from a Wikipedia page about chemistry, asking what year the term alkaloid was introduced.
In this sample, the image is of Naloxone, which is an alkaloid, and the table is about alkaloids, but the answer is actually in the text. Furthermore, the question is ambiguous because it could easily be found in a table or a paragraph. The agent must reason over each modality to determine where the answer can be found before producing an answer.
Figure \ref{fig:example2} (see next page) features a question from a page about sign language, asking the country from which the manual alphabet and numbers come from. Each modality in this example is about American Sign Language (ASL); the image contains various hand-signs associated with letters, the table is about a hand-sign, and the text discusses ASL. Again, the agent must first reason over these modalities before it can produce an answer, locating it in either the image caption or the text.

\end{document}